\documentclass{article}
\usepackage{ijcai20}

\usepackage{times}

\usepackage{soul}
\usepackage{url}
\usepackage[hidelinks]{hyperref}
\usepackage[utf8]{inputenc}
\usepackage[small]{caption}
\usepackage{graphicx}
\usepackage{amsmath}
\usepackage{booktabs}
\usepackage{lipsum}
\usepackage{xcolor} 

\usepackage{amssymb}
\usepackage{amsfonts}
\usepackage[ruled,vlined]{algorithm2e}
\SetKwInput{KwInput}{Input}                
\SetKwInput{KwOutput}{Output}              

\usepackage{array}
\usepackage{multirow}

\usepackage{booktabs}
\usepackage{subcaption}
\usepackage{pifont}
\usepackage{color}
\usepackage{textcomp}

\hypersetup{colorlinks,linkcolor={blue},citecolor={blue},urlcolor={red}}  

\title{Recent Advances in Understanding Adversarial Robustness of Deep Neural Networks}

\author{
Tao Bai$^1$\footnote{Contact Author}\and
Jinqi Luo$^1$\and
Jun Zhao$^{1}$\\
\affiliations
$^1$Nanyang Technological University, Singapore\\
\emails
\{bait0002, luoj0021, junzhao\}@ntu.edu.sg
}

\begin{document}
\maketitle
\begin{abstract}
Adversarial examples are inevitable on the road of pervasive applications of deep neural networks (DNN). Imperceptible perturbations applied on natural samples can lead DNN-based classifiers to output wrong prediction with fair confidence score. It is increasingly important to obtain models with high robustness that are resistant to adversarial examples. In this paper, we survey recent advances in how to understand such intriguing property, i.e. adversarial robustness, from different perspectives. We give preliminary definitions on what adversarial attacks and robustness are. After that, we study frequently-used benchmarks and mention theoretically-proved bounds for adversarial robustness. We then provide an overview on analyzing correlations among adversarial robustness and other critical indicators of DNN models. Lastly, we introduce recent arguments on potential costs of adversarial training which have attracted wide attention from the research community.
\end{abstract}


\section{Introduction}
The adversarial vulnerability of deep neural networks has attracted significant attention in recent years. 
With slightly crafted perturbations, the perturbed natural images, or Adversarial Examples~\cite{DBLP:journals/corr/SzegedyZSBEGF13} can mislead state-of-the-art classifiers to output erroneous predictions.
Other than classification, previous works has also proved the existence of adversarial examples in varieties of computer vision tasks: Semantic Segmentation~\cite{DBLP:conf/iccv/XieWZZXY17}, Object Detection~\cite{ijcai2019-134,jia2020fooling}.   Super-Resolution~\cite{Choi_2019_ICCV} and in other fields~\cite{alzantot2018generating,carlini2018audio,huang2017adversarial,dai2018adversarial}, which arises concerns of the public, especially in safety-critical areas.

As extensive methods of generating adversarial examples appear~\cite{DBLP:journals/corr/GoodfellowSS14,carlini2017towards,DBLP:conf/iclr/MadryMSTV18}, the countermeasures are proposed at the same time~\cite{DBLP:journals/corr/abs-1711-00117,DBLP:conf/cvpr/LiaoLDPH018,DBLP:conf/cvpr/LiaoLDPH018,balunovic2020adversarial}, which seems like an arm race.
It is common to see that an defense is defeated by an newer attack within a short time, and vice versa.
However, this arm race does not help us much to understand adversarial examples.
The reasons for adversarial examples' existence remains unclear and it is still intractable to produce adversarially robust deep neural networks.
Currently adversarial training~\cite{DBLP:journals/corr/GoodfellowSS14,DBLP:conf/iclr/MadryMSTV18} is recognized the most effective way to gain adversarial robustness in practice, where the neural networks are forced to play a \textit{min-max} game.
Despite its effectiveness, it has shortages like computationally intensive, time-consuming, and non-provable. 

To improve the reliability and trustworthiness of deep neural networks intrinsically, in the last few years, more and more researchers begin to focus on analyzing and explaining deep neural networks in the adversarial setting from different perspectives~\cite{DBLP:journals/corr/GoodfellowSS14,pang2020rethinking,geirhos2018imagenettrained,DBLP:conf/icml/ZhangZ19,NIPS2019_8307,NIPS2019_9394,Wang_2020_CVPR}.
With extensive efforts of the whole community, it is getting clear that why deep neural networks are sensitive to imperceptible perturbations, and some clues are found related to adversarial robustness.

Some recent surveys on adversarial examples have been proposed~\cite{Akhtar2018,wiyatno2019adversarial}, which give comprehensive discussions on existing adversarial attack and defense methods.
To our best knowledge, surveys focusing on understanding adversarial robustness do not exist even though the past few years have witnessed the great efforts made by this community. 
We therefore believe that this survey on adversarial robustness can provide up-to-date findings and developments happening in this field. 

Our goal in this paper is to give a brief overview of adversarial robustness.
Particularly, we carefully review and analyse adversarial robustness from different perspectives, uniquely highlight the challenges and present future research directions.

\section{Preliminaries}
\subsection{Adversarial attacks}
Adversarial attacks usually refer to finding adversarial examples for well-trained models.
Formally, finding adversarial examples is like solving a optimization problem.
We denote by $f(x, \theta): \mathbb{R}^{m} \rightarrow\{1 \dots k\}$ an image classifier that maps an input image $x$ to a discrete label set $C$ with $k$ classes, in which $\theta$ indicates the parameters of $f$.
Then the adversary tries to find a perturbation $\delta$ as follows:
\begin{equation}\begin{array}{c}
\Delta^*:=\mathop{\arg\min}\limits_{\delta} \{ \delta\in \mathbb{R}^{m}  \mid |\delta|_p \leq \epsilon, f(x+\delta)\neq f(x) \},
\end{array}\end{equation}
where $\epsilon$ is to bound the magnitude of $\delta$, and $p$ can be 0, 2 and $\infty$. 
In most cases, $\epsilon$ tends to be small so that the perturbations are imperceptible to human eyes.


\subsection{Adversarial Robustness}
Currently adversarial robustness is defined as the performances of well-trained models facing adversarial examples.
In practice, the widely accepted way to gain adversarial robustness is adversarial training~\cite{DBLP:conf/iclr/MadryMSTV18}.
The idea of adversarial training is simple: it requires targeted models to train on adversarial examples in each training loops.
Formally, the objective function of adversarial training is 
\begin{equation}\begin{array}{c}
\mathcal{L}(\theta, x, y):=\mathbb{E}_{(x, y) \sim D}\left[\max \limits_{\delta \in B(x,\varepsilon)} \mathcal{L}(\theta, x+\delta, y)\right]
\label{eqn:adv train}
\end{array}\end{equation}
where $B(x,\varepsilon) := \left\{x+\delta \in \mathbb{R}^{m}  \mid  \ \mid \delta\ \mid _{p} \leq \varepsilon\right\}$, and $\delta$ is obtained via gradient-based methods discussed above.
For most of time, the method used to maximize $\mathcal{L}(\theta, x+\delta, y)$ is Projected Gradient Descent~(PGD) which calculates the gradients iteratively:
\begin{equation}
\delta^{t+1}:=\Pi_{B(x,\varepsilon)}\left(x + \delta^{t}+\alpha \operatorname{sgn}\left(\nabla_{x} \mathcal{L}(\theta, x+\delta^t, y)\right)\right)-x
\end{equation}
where $\Pi_{B(x,\varepsilon)}(\cdot)$ is a projection operator projecting the input into the feasible space $B(x,\varepsilon)$, $t$ is current step, and $\alpha$ is the step size.
Observed from Eq.~(\ref{eqn:adv train}), adversarial training encourages models to predict correctly in an \mbox{$\epsilon$-ball} surrounding data points. Many variants of adversarial training have been developed from this observation~\cite{Zhang2019TheoreticallyPT,jin2020manifold}.

\section{Understanding Adversarial Robustness}

\subsection{Robustness Evaluation}
In this section, we introduce recent works on evaluation of adversarial robustness from different perspectives.

\subsubsection{Benchmarks}
Model accuracy to adversarial examples generated by attacks (FGSM, I-FGSM, FGS, rFGS, DeepFool, CW \cite{7958570}) is the direct indicator of adversarial robustness. \cite{tramer2020adaptive,croce2020reliable} show the insufficiency of such straightforward adaptive approach. \cite{croce2020robustbench} try to establish a systematic standardized benchmark for adversarial robustness by imposing restrictions on the defenses involved.

Additionally, radius of distortion represents the range of adversarial perturbation for generating successful Adversarial Examples. The magnitude of this variant is also meaningful for evaluation of model robustness. With all other variants unchanged, an adversarially-trained model that is robust to AE of larger perturbation radius usually suggests that it has higher adversarial robustness.

\cite{Mu2019MNISTCAR} introduced the MNIST-C dataset by applying visual corruptions on the MNIST testset for benchmarking out-of-distribution robustness. \cite{vaishnavi2019attention} provided two datasets with foreground attention masks that are generated from The German Traffic Sign Recognition (GTSRB) \cite{germanroadsigndata} and MS-COCO to facilitate adversarial training.

\cite{hendrycks2018benchmarking} introduced a series of widely-used visual corruptions and adversarial perturbations to generate a dataset (ImageNet-C)  for corruption robustness and a dataset (ImageNet-P) for adversarial perturbation robustness. \cite{Hendrycks2019NaturalAE} further proposed a dataset (ImageNet-A) which includes 7500 naturally occurring adversarial examples that are encapsulated as a testset for ImageNet-based classifier. After this, in \cite{Hendrycks2020TheMF}, authors introduced a visual-art-based dataset (ImageNet-R) containing renditions of 200 ImageNet classes.

 Other than static images, \cite{Gu2019UsingVT} pioneered in open-sourcing datasets for evaluating robustness to the minute transformations across video frames. Later in \cite{Shankar2019DoIC}, ImageNet-Vid-Robust and YTBB-Robust were proposed to test the robustness on temporal perturbations derived from animated videos.

\subsubsection{Theoretical Bounding}
Pioneering work \cite{Fawzi2015AnalysisOC} has provided fundamental upper bounds and an evaluation metrics for analyzing the robustness of classifiers to adversarial attacks based on the complexity of classification tasks. \cite{DBLP:journals/corr/abs-1808-01688} introduced Accuracy-Robustness Pareto Frontier (ARPF) for deep vision models to evaluate robustness towards various adversarial attacks. \cite{buzhinsky2020metrics} proposed to evaluate the robustness with the help of generative models. Within this work, authors uses latent space performance metrics, which is based on probabilistic reasoning in latent spaces of generative models, to measures the model's resistance to adversarial examples. \cite{NIPS2019_8968} used optimal transport to study the gap between the possible optimal classification accuracy and the currently achieved SOTA ones by adversarially-trained neural networks.

\cite{weng2018evaluating} proposed Cross-Lipschitz Extreme Value for Network Robustness (CLEVER), which is a Extreme Value Theory (EVT) based metric for measuring the robustness of DNN without evaluating adversarial attacking types. However, \cite{Goodfellow2018GradientMC} pointed out that gradient masking 
can misguide CLEVER to overestimate the magnitude of perturbation needed to fool the model. This makes CLEVER an inaccurate evaluation of adversarial robustness. Towards improving CLEVER, \cite{DBLP:journals/corr/abs-1810-08640} provided two directions. In the first part, EVT is applied on a new twice-differentiable robustness guarantee to imply a new estimation on robustness score, namely the second-order CLEVER score. Then authors involved Backward Pass Differentiable Approximation (BPDA) to equip CLEVER, showing its broader application on DNN with non-differentiable input transformations.

\subsection{Correlation Analysis}
Adversarial robustness of DNN can be intrinsically related to many properties of the model and datasets. Models with different levels of robustness can also have different bias towards features. In this section, we list several possible directions of research on relations of adversarial robustness and model property, dataset distribution, and training settings.

\subsubsection{Feature Representation}
\cite{NIPS2019_8307} divided feature representations into two categories, namely Robust Features and Non-Robust Features, by a purely human-centric approach. Robust features are features correlated to the object label that can be recognized by human even after generation of AE. Non-robust features are the brittle and incomprehensible features from patterns that can be flipped with adversarial attack. Authors claim that adversarial vulnerability is a direct result of sensitivity to well-generalizing features in the data. By proposing an theoretical framework and experimenting on adversarial attacks, they defend their hypothesis that AEs are a natural consequence of the presence of highly predictive but non-robust features. \cite{vaishnavi2019attention} showed that utilizing foreground attention masks on training sets, which addresses the object's foreground features and eliminates the effect of noise background, can have positive contribution towards increasing PGD-based adversarial robustness.

\cite{pmlr-v97-etmann19a} quantified the relation between model robustness and interpretability of input's saliency map, finding out that the adversarial robustness and alignment of salience map is positively related when testing on non-complex datasets like MNIST. They proved, both theoretically and experimentally, that DNN models that are more robust can produce saliency maps with clearer indication of outstanding features. \cite{ding2018on} discovered the relationship between adversarial robustness and the intrinsic distribution of training data. They hypothesis that alternation on the data distribution that is semantically stable could have significantly different outcome on robustness of adversarially-trained neural networks. Authors also pioneered in raising attentions on developing proper evaluation of adversarial robustness since existing datasets cannot be suffiently reliable after semantically-loseless shift. \cite{pmlr-v97-simon-gabriel19a} found out that, when training loss converges around its minimum, the test accuracy can still be slightly improved but such increment will severely threaten DNN’s robustness because the last increment on accuracy demands a significant increase on loss gradients.

\subsubsection{Model Sparsity}
DNN are computationally expensive and therefore techniques of model pruning have been widely used to maintain the model sparsity. \cite{NIPS2018_7308} conducted research on classification tasks of DNN model to show that, both theoretically and experimentally, the model sparsity and its adversarial robustness has intrinsic relations. The authors find out that the training process of non-linear DNN model can achieve higher adversarial robustness when trainers can increase model sparsity properly through pruning techniques. However, sparse DNN that are over-pruned can be more vulnerable to adversarial attacks than original dense DNN.

\cite{rice2020overfitting} showed that overfitting training in the adversarial context will harm the model robustness. Their empirical results indicate that early stopping will help to reserve the robust performance. Similarly, \cite{Ma_2020} found the the state-of-the-art DNNs designed for large-scale image datasets are often over-parameterized on medical imaging tasks. Their experiments show that the lacking of model sparsity leads to low adversarial robustness of medical-image-based DNNs towards adversarial examples.

\subsubsection{Model Bias}
\cite{DBLP:conf/icml/ZhangZ19} attempted to discover the inherent bias of Adversarial-Trained Convolutional Neural Networks (AT-CNN) during the task of image recognition. By experimenting on salience maps and alternations of ImageNet destroying either texture or shape, authors find that AT-CNN, which has higher robustness on adversarial attacks, are more biased towards learning global structures (shapes, edges) when recognizing an object. Oppositely, normally-trained CNN with lower robustness demonstrate different bias such that it is more favorable of recognizing the object's texture information. \cite{geirhos2018imagenettrained} drew a similar conclusion stating that, although CNN are widely considered to be shape-biased when recognizing objects, their experiment results on ImageNet-trained CNN show the phenomenon that CNN models are instead strongly biased towards learning the representation of textures rather than shapes. They also find that CNN trained on ImageNet with texture information removed are of higher robustness on classification accuracy when adding various noises to the input image.

\subsubsection{Local Smoothness}
\cite{Yu2019InterpretingAE} found that adversarially-trained DNN models are shown to have more smooth and stable geometry at the peak of loss surface (the visualization of multi-dimension loss function). Such smoothness relieves the under-fitting phenomena on adversarial examples lying in the neighborhood of inputs. \cite{Fawzi2015AnalysisOC} suggested that the phenomenon of adversarial vulnerability is partially because of the low flexibility of classifiers around the varying decision boundary when the difficulty of the classification task changes. \cite{Yu2019InterpretingAE} appliec a visualization approach named decision surface to investigate adversarial robustness. Their approach confirms the intuition that adversarial examples are naturally-existed data points lying in the close neighborhood of the given inputs.

\subsubsection{Corruption Robustness}
\cite{pmlr-v97-gilmer19a} stated that adversarial robustness and corruption robustness are two manifestations of the same underlying phenomenon. Their approach builds up close connections between two concepts by both theoretical proof and empirical experiments. The result suggests that the training process of adversarially-robust DNN model should take improving performance on realistic image corruptions (e.g. ImageNet-C) as an important factor.

\subsection{Underlying Cost}
Enhancing adversarial robustness is not always favorable when taking more performance indicators into consideration. In this section, we will introduce a few widely-concerned dilemma relevant to DNN robustness.
\subsubsection{Adversarial Robustness against Accuracy to Natural Examples}
Despite the success in improving robustness of neural networks to adversarial attacks, adversarially trained models frequently fails to generalize on natural test samples. In this section, we summarize discussions on trade-off between natural accuracy and adversarial robustness. 

Early works \cite{Rozsa_2016} showed that normal models with higher accuracy appears to be more robust on adversarial examples. However, authors of \cite{hendrycks2018benchmarking} discovered that more accurate models are actually less robust since their seemingly higher robustness is because of the increased natural accuracy. \cite{Balaji2019InstanceAA} proposed instance-adaptive adversarial training that manipulates the perturbation radius of every training sample to fit a more generalized model with higher natural accuracy. \cite{buzhinsky2020metrics} showed that there is a correlation between the adversarially trained DNN accuracy on clean images and its adversarial robustness in latent spaces. \cite{DBLP:journals/corr/abs-1808-01688} evaluated the recent state-of-art ImageNet-based DNN models on multiple robustness metrics. Authors figured out the linearly negative correlation between the logarithm of model classification accuracy and model robustness. Theoretical analysis on their correlation was also provided in \cite{Zhang2019TheoreticallyPT,tsipras2018robustness}.

More recently, the long standing trade-off between adversarial robustness and natural accuracy is shown to be not necessary unavoidable \cite{Stutz2019DisentanglingAR,yang2020closer}. \citeauthor{yang2020closer} argued that image datasets including CIFAR-10 and MNIST are shown to be distributionally separated and this separation implies that the existence of a robust and accurate classifier can be obtained by rounding a locally-Lipschitz function. Authors also demonstrated that using dropout can facilitate the generalization of adversarial training. Similarly, \cite{song2018improving} showed that such generalization gap of robust training can be narrowed by domain adaptation.

\subsubsection{Adversarial Robustness against Training Resources}
The SOTA adversarial training approaches boast for their high robustness. However, such training process on huge datasets like ImageNet has been only accessible in limited laboratories around the world due to high computing cost, which is partially caused by the high sample complexity during adversarial training \cite{NIPS2018_7749}.

In \cite{Shafahi2019AdversarialTF} authors proposed FREE, an efficient adversarial training algorithm that enables a much cheaper cost of time on adversarial training compared to traditional approaches. Authors argue that it is more efficient to update both the model parameters and image perturbations simultaneously in one backward pass rather than updating them separately. Further after FREE, in \cite{Wong2020FastIB}, authors proposed FAST that introduces random initialization points to make FGSM-based adversarial training as effective as the projected gradient descent based training. They also argue that most of the popular adversarial training approaches can be accelerated by applying standard techniques of efficient training for normal deep networks. However, \cite{Andriushchenko2020UnderstandingAI} argued that adding such unnecessary randomness in FAST does not prevent catastrophic over-fitting where the model drastically loses robustness on a few epoch of adversarial training. Alternatively, authors proposed GradAlign, a new regularization method that prevents catastrophic overfitting by explicitly maximizing the gradient alignment inside the perturbation set. From the perspective of Pontryagin’s Maximum Principle (PMP), authors of \cite{Zhang2019YouOP} discovered that the adversarial perturbation is mostly coupled with the weights of the first layer of the neural networks. This work introduced YOPO, an Neural Network-oriented adversarial training algorithm that splits up the adversary computation and weight updating, where the adversary computation is only focused on the first layer of the proposed network.

\section{Conclusion}
Training robust models that are secure and reliable has become a necessary step towards trustworthy applications of deep learning. In this short survey, we provide an overview on adversarial robustness and present recent approaches of robustness evaluation. After that, we study recent advances in how to analyze and understand underlying properties of adversarial robustness. Finally, we summarize the existing discussions on potential costs of improving adversarial robustness. 
\bibliographystyle{named}
\bibliography{ijcai20}

\end{document}